# A field study on Polish customers' attitude towards a service robot in a cafe

Maria Kiraga[1]*, Zofia Samsel[2]* and Bipin Indurkhya[3]

[1] AGH University of Science and Technology, Kraków, Poland
marysiakiraga@gmail.com
[2]Learning Planet Institute, Université Paris Cité, Paris, France
zosiasamsel@gmail.com
[3] Cognitive Science Department, Jagiellonian University, Kraków, Poland
bipin.indurkhya@uj.edu.pl
* These authors contributed equally.

**Abstract.** More and more stores in Poland are adopting robots as customer assistants or promotional tools. However, customer attitudes to such novelty remain unexplored. This study focused on the role of social robots in self-service cafes. This domain has not been explored in Poland before, and there is not much research in other countries as well. We conducted a field study in two cafes with a teleoperated robot Nao, which sat next to the counter serving as an assistant to a human barista. We observed customer behavior, conducted semi-structured interviews and questionnaires with the customers. The results show that Polish customers are neutral and insecure about robots. However, they do not exhibit a total dislike of these technologies. We considered three stages of the interaction and identified features of each stage that need to be designed carefully to yield user satisfaction.

## 1    Introduction

The service industry is increasingly accepting the use of robots to help with repetitive or physically demanding tasks, but also as active assistants directly interacting with customers. In the past, such interactions and customer service in restaurants or cafes were carried out exclusively by humans. Now it is useful to explore how to deploy robots as sales and service assistants to reduce human service duties and speed up customer service time. However, introducing new technologies in everyday life is received differently in different cultures and religions [1]. Hence, culture-centered research on customers' attitudes and expectations toward robot assistants is needed to design an interaction scheme to benefit both the workers and the customers.

We investigated Polish consumers' attitudes and behaviors towards a robotic assistant to determine which interactions lead to a positive customer experience. A field study was conducted with a teleoperated social robot Nao to recommend products and take orders from customers. We observed customer perceptions of robot assistants in cafes to study the best way to deploy them. Though Polish customers' attitudes toward robots vary widely, we distinguish three aspects of the interactions that influenced people's attitudes and experience: the first impression (robot's physical appearance and behavior), the flow of conversation, and contextual adaptation. Our research complements state of the art on human-robot interaction (HRI) in terms of interactions with service robots and individualized approaches to creating such interactions.

## 2 Related Research

### 2.1 Attitude of Polish Society Towards Technology and Robots

Poland has a rapidly developing economy [2]. Its entrepreneurs are eagerly seeking new technological solutions to streamline and secure their work. Based on a 2022 report, 97% of Poles believes that new technologies are generally important and useful, however over half of the participants were sceptic toward the use of technologies on a daily basis [3]. On the other hand, one study in which the researchers explored the attitudes of elderly Poles towards the utilization and ease of use of domestic robots showed that they were eager to learn and adapt to modern devices [4].

There is no existing research on the general attitude of Poles toward robots. But it is possible to characterize Polish society through the perspective of its history, religion, and culture, which are significant variables that differentiate attitudes toward technologies and the use of robots in public places [5]. Based on the Polish Statistics Census in 2021, approximately 85% of Poles described themselves as Catholics [6]. This, and the fact that Christians seem to be skeptical about the use of robots in everyday life [7], we can expect that the Polish public may have a negative attitude towards robots in a public place. Also, traditionalism and the high importance attached to family values [8] may negatively influence their approach to robots.

### 2.2 Use of Robots in Public Spaces in Poland

Current reports of "robotization" of Poland mainly concern industrial robots, portraying Poland as robustly developing in this area [9]. Though robots are not commonly used in public places, their numbers in the sales and service section are increasing, offering a potential to automate more social jobs [10]. *Kerfuś* - a cat-robot used in Carrefour supermarket [11] gained popularity among social media users shortly after being introduced in 2022: this robot has been deployed at a few Shell gas stations since 2021. A robot *BellaBot* delivers ordered items from a catering offer directly to the table, while another, *KittyBot,* informs about the current promotional campaigns [12]. Restaurants in Poland are also finding uses for robots. An autonomous robot waiter transporting food to tables was used at a Hilton hotel restaurant in 2021 [13].

All these robots are autonomous robots that do not have extensive conversation or order-taking capabilities. According to the literature, customers and visitors perceive more positively the interaction with a robot assistant in a public space that has broad capabilities to communicate and adapt to the situation [14].

### 2.3 Robot Design in Human-Robot Interaction

Past research suggests some significant features affecting a customer's perception (positive or negative) of the robot. In the services sector, it is necessary for the robot to have specialized knowledge and communication skills. Trust plays a crucial role in influencing acceptance of the robot [15]. Customers expect robots to conform to social norms, while not having to reciprocate politeness; they are expected to behave in a humanlike way [14]. This can be provided by teleoperated robots where a human operator controls all the robot's behavior, gestures, and speech: such robots have been found to be successful [16]. Furthermore, the teleoperator observes the robot environment and can provide relevant advertising strategies depending on the context [17]. However, robots that are observed to be autonomously controlled can reduce assessment concerns and increase customer confidence [16].

For social robots in customer service, just verbal communication with customers is not sufficient. If robots are animated and have human-like features, they are viewed as more approachable and capable, which leads to increased customer satisfaction [18], [19]. It has been shown that the pressure to not say 'no' to a human salesman is higher than to a robot, so interaction with a robot can be more relaxing for a user [20]. Though deploying robots as store assistants can enhance efficiency for both the customer and the organization, many people still prefer communicating with another human [21]. One way to compromise is using a humanoid robot to assist customers in situations where human staff needs to focus on more complex tasks [22].

While robots can be fully operational and serve a specific purpose, users may still feel mistrust, insecurity, and anxiety around the robot [23, 15]. Therefore, it is important that the operator, especially in the case of a salesperson robot, manages the behavior correctly to ensure the best possible customer experience.

### 2.4 Choice of Robot and the Wizard-of-Oz Methodology

We used the humanoid robot Nao (from Softbank Robotics), which was remotely teleoperated by a human [24]. We conducted seven field study sessions using this Wizard-of-Oz methodology. The participants interacting with the robot were made to believe that the robot is autonomous. This allowed for naturalistic settings, which is the main idea behind conducting field studies without having to face possible errors and mistakes when the robot is running by a pre-programmed artificial intelligence.

## 3 Session One

### 3.1 Method

This session took place in *Pastelove* – a cafe in one of Jagiellonian University's departments. Nao was placed at the entrance, where the queue forms. The teleoperator was at a table next to Nao (Fig. 1). Nao tried to make small talk with the customers waiting in line and promote certain products.

The session lasted for two hours with five people getting into a conversation with the robot. We estimate that about 50 people visited the cafe in this time. The data collected included our observations, transcripts of the customers' interactions with Nao, and transcripts of our short, semi-structured interviews with some customers who agreed to take part. The questions asked included: satisfaction after the interaction, likeliness of getting into a similar interaction in the future, and attitudes towards similar robots. All the robot's questions and answers were typed into a text-to-speech engine by the experimenter. The questions were written beforehand and selected using button, but the answers were improvised on the spot.

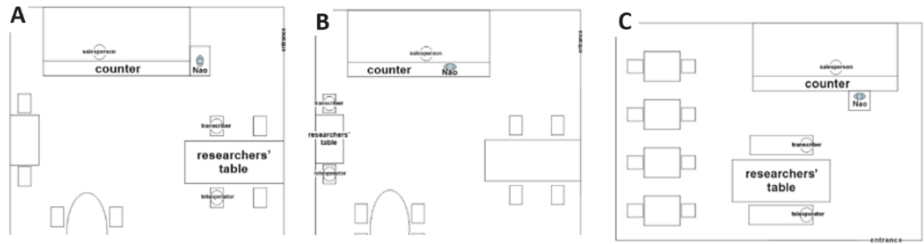

**Fig. 1.** Setups for the first session in *Pastelove* (A) and sessions two to seven in *Patelove* (B) and *Karma* (C).

### 3.2 Results and discussion

All the data was qualitatively analyzed. Only some visitors interacted with the robot: most of them were only interested in taking pictures of the robot or greeting it briefly. Various robot's prompts such as "*Hey, how are you?*" or "*Do you like gingerbread spiced coffee?*" were ignored or met with comments to other customers, which revealed fear of the robot and surprise at its presence. Five people who talked to the robot mainly made small talk on various topics (e.g., weather, coffee preferences).

In the interviews, the customers expressed positive emotions towards the interaction. The appearance of the robot was mostly commented on in a positive manner. The flow of conversation was perceived to be intelligent. In addition, they expressed surprise with such technology being present in a university cafe, as none of these five participants had previous experience with similar robots. One customer noted that "*talking to a robot was slightly weird*", because it reminded them of chatting with a child. (Nao robot is small and has a humanoid appearance.)

After this session, we made some adjustments to provoke more interactions from the customers. Previous research has shown that the closer a person gets to a robot, the more trustworthy the robot is perceived to be, which facilitates a positive interaction [25]. Moreover, Nao making only small talk does not seem to be a strong enough cue for the customers to start interacting with such, supposedly new for them, technology. Also, its appearance does not indicate its role as a staff member: it seems to be a toy or a mascot, which makes prompting the interaction more difficult. From other studies we learn that the robot partner's initiative leads to increasing the trust of the human partner [16], which suggests the robot's approach should be more proactive.

## 4 Sessions Two to Four

### 4.1 Method

With the results of the previous session in mind, we changed the setup and the role of the robot. These sessions were conducted over three days, two hours each day. One session took place in the same cafe as the pilot study (*Pastelove*), and the following two sessions took place in a small cafe *Karma*, where the robot was placed right by the counter. The setup in *Karma* is shown in Fig. 1. The teleoperator sat at a table across the counter facing away from the customers. The researcher making observations sat next to the teleoperator. Participants were customers of the cafes. Most of the customers were not aware of the robot's presence before coming to the cafe.

This time, instead of simply making small talk, Nao was serving as an assistant to the baristas. Its role was to get an order from the customer, take it back to the human salesperson, and chat with the customer. The flow of interaction was designed beforehand and is shown in Table 1. To make the interaction more natural, some answers, which we did not include in the interaction flow, were improvised on the spot.

As customers entered the cafe and approached the counter, Nao welcomed them and started the ordering procedure. The language of interaction was either English or Polish depending on the customer. If a customer was eager for more interaction, the teleoperator improvised to make the interaction as natural as possible, personalizing the robot's conversation to each customer. If a customer was disinclined to interact with Nao, they could place the order directly with the human barista behind the counter. After the interaction was finished, a researcher approached the customers to request an interview. Both the customers who engaged with Nao, and those who did not, were asked for a short interview. A total of 31 customers agreed to take part.

As before, we focused on qualitative data. (We did not count the exact number of people visiting the cafes, or how many chose to have the conversation and how many did not.) The data collected included our observations, transcripts of the customers' interactions with Nao, and transcripts of our short, semi-structured interviews with the customers. All the data was qualitatively analyzed for extracting those features of the interactions that mattered the most for the customers.

**Table 1.** Flow of interaction for the cafe sessions

| |
|---|
| **Stage 1: Placing an order** |
| After each product is ordered, the robot asks the customer: *"Would you like anything else?"* and recommends some product (e.g., *"We have tasty coffee/croissants."*). |
| The robot asks if the order is to be prepared here or to-go (e.g., *"Eat in or takeaway?"*) |
| **Stage 2: Transferring the order to the barista** |
| The robot repeats the order and thanks you for placing it. An example phrase used was: "*Thanks for placing the order! It will be ready in a minute!*" |
| **Stage 3: Small talk** |
| After receiving the order, the robot asks the customer if he or she would like to talk to it while waiting: (e.g., *"While you're waiting, would you like to talk with me?"*) |

## 4.2 Results and discussion

Placing the robot on the counter and changing the topic of the interaction had a positive impact and increased the number of people who approached Nao. The new setup required them to get close to the robot so they could not ignore it completely. However, because we did not count the number of people in the cafe, we cannot conclusively say whether this was due to better design of interaction, the placement of the robot, or because more people came to the cafe.

We observed three aspects of human-robot interaction that determined the progress of the interaction and perception of the robot: people's *first impressions* of the robot, including their opinions on the robot's physical appearance and behavior, the *conversation flow*, and the robot's *adaptation to specific contexts*.

**First impressions.** Appearance, the voice, and speech pattern of the robot seemed to be the first elements that determined whether a person would be willing to initiate an interaction with the robot or not. Here we distinguished behaviors related to the positive perception of the robot's appearance: the most common positive reaction to the robot's appearance was that *"it's cute", "his eyes are glowing", "wow, he's sitting like a king of this cafe"*, which are related to anthropomorphizing and the novelty of this technology. Some people were surprised with Nao following them with its eyes ("*It looks at me and knows where I am.*") or even by talking to them directly and with cohesion.

However, during this initial phase, many people refused to participate in the interaction with Nao. Many such people expressed their dislike of the robot appearance by criticizing its appearance: *"It looks scary";* or being afraid or disgusted by the robot's way of speaking, summed up with a comment: "*The child's voice seems frightening, like there's something trapped inside the robot*". These people usually avoided looking at the robot or noting its presence explicitly. They stood far away from the robot while making the order.

Some of these people also expressed their uncertainties about the robot's intelligence and abilities. For example, whether the robot would understand them: "*Will he know what I am saying?*". One person asked, "*Can I take a picture of the robot? I don't have to ask him, right? He's just a machine.*" One customer said to the barista that the robot "*lacks a lot to run such a business*", and "*the robot won't be able to replace human assistants in cafes*". One man who wanted to buy coffee beans ignored the robot and asked the barista for advice on possible choices. One woman ignored the robot when she wanted to know what food was currently available to order. Some customers did not know how to behave and how they should approach the robot. Some people seemed interested in the robot's presence in the cafe, but commented on its behavior to the barista, ignoring what the robot was saying.

**Conversation flow.** People who decided to take part in the interaction reacted differently to the robot's voice and its way of speaking. Some of them described it as "*talking to a pet*" or like chatting with a child. They did not focus on mistakes made by the robot while speaking, or long pauses in robot responses. Often, they initiated the interaction and asked the robot questions unrelated to customer service.

However, there were some customers who were willing to start the interaction but got confused or irritated when the robot did not respond quickly enough. Some people gave up on the interaction and went straight to the barista to make the order. Often, a customer said: "*it's talking too slow!*" and dropped out of further conversation.

**Adaptation to specific contexts.** Some customers who ordered from the robot were interested in continuing further conversation. (In contrast, some others passively answered questions asked by the robot and did not want to continue the conversation after placing an order.) We observed specific behaviors of the robot, which facilitated the willingness to interact with it. The first of these was telling jokes or stories. Another aspect was when the robot addressed them directly and complimented their clothing or other accessories. Additionally, continuous interaction was encouraged by the robot addressing the customer by name.

As the conversation became longer, people started to comment on the robot's intelligence. They referred to it as "*smarty-pants*" or that it is generally smart.

In general, many factors affect the success of carrying out the interaction. The most important moment in the interaction seems to be its initiation - the first impression of the robot and interaction. As this was the stage when most people decide whether to participate in the conversation, we aimed to determine the customers' approach toward the robot. To check more precisely the attitudes and expectations of customers towards the robot and its functions in the cafe, we conducted additional sessions.

## 5 Sessions five to seven

### 5.1 Method

These sessions were conducted over three days in the cafe *Karma*. The study setup and interaction design were the same as in sessions two to four. During these sessions we collected data on the number of visitors to the cafe ($N = 67$).

Additionally, for gathering specific data on the customers' attitudes towards robots we adapted a questionnaire to our cafe setting [26]. Visitors of the cafe were approached by the researchers after placing an order (to either the human or the robot) with a request to fill in the questionnaire. Some of them ($N = 26$) agreed to respond.

The questionnaire (Table 2) was divided into five sections: *Interest*, *Negative attitude*, *Utility, Appearance* and *Familiarity*. The first three gave us information about the general attitude of the customer towards robots as well as the perceived potential of service robots in cafes. The latter two sections provided insights on desired physical features of the robot and customer's previous experience with robots. Participants were asked to state how strongly they agree with given sentences on a scale 1 ("*strongly disagree*") to 5 ("*strongly agree*"). We also gathered the basic demographic information about participants' age, gender and language used.

In this session, we did not record observations of the customers' behavior, but analyzed the participants' (*N = 9*) responses to the post-interaction interview, and the questionnaire data from (*N = 26*) participants.

### 5.2 Results and discussion

**First impressions:** The customers displayed similar behaviors towards the robot as in the previous sessions. Among the nine people who interacted with the robot by giving their order to it, four made comments about the robot's pleasant appearance and voice being an advantage. On the other hand, to the question "*What didn't you like about the interaction?*" one person responded: "*The fact that Nao was stiff and did not make free movements while talking*".

For Nao's conversation approach, three participants appreciated Nao's politeness, with one of them commenting: "*Overall, the whole experience was unique, the robot was very polite*".

Three of the nine people have had previous experience in talking to similar robots. The remaining six people reflected on their discomfort connected to their first human-robot interaction principally focusing on "*an awkward feeling*".

**Table 2.** Categories of interaction for the café session

| Categories of interaction for the café session | *M* | *SD* |
|---|---|---|
| **Interest** | | |
| I think a robot can be a communication partner. | 3.46 | 1.24 |
| I want to converse with a robot. | 3.12 | 1.42 |
| I would want to boast that I have interacted with a robot. | 2.77 | 1.48 |
| It is good if a robot can do the work of a human. | 3.46 | 1.30 |
| I feel at ease around robots because I do not need to pay attention to robots as I do to humans. | 2.58 | 1.50 |
| **Negative attitude** | | |
| It would be a pity to have a robot in my favorite cafe. | 3.00 | 1.60 |
| The movements of a robot are unpleasant. | 1.92 | 1.26 |
| It is unnatural for a robot to speak in a human language. | 1.50 | 0.86 |
| I feel like I also become a machine when I am with a robot. | 1.38 | 0.90 |
| I feel scared around robots. | 4.52 | 1.04 |
| **Utility** | | |
| I think a robot could make food/coffee recommendations to me. | 3.23 | 1.24 |
| I think a robot would understand my order. | 1.88 | 1.10 |
| I like the idea of using robots in cafes or stores as assistants. | 2.92 | 1.32 |
| I like the idea of using robots in cafes or shops to entertain customers. | 2.38 | 1.27 |
| **Appearance** | | |
| I think the robot design should be cute. | 3.69 | 1.38 |
| I think robots should have animal-like shapes. | 3.69 | 1.38 |
| I think the voice of a robot should be like the voice of a living creature. | 3.31 | 1.40 |
| I think a robot should have human-like shape. | 3.92 | 1.13 |
| **Familiarity** | 1.39 | 0.64 |
| I often come across robots in public spaces. | | |
| I would be less surprised to come across a robot in public space in country other than Poland. | 3.06 | 1.31 |

**Conversation flow:** The main disadvantage turned out to be that the interaction was not fluent. Five of the nine people complained about the answers being too slow, which caused the interaction to become more mechanical and less natural.

**Adaptation to specific contexts:** This aspect is as crucial to the customers as in the previous sessions. The conversations were often judged as intelligent. Nao's answers to the customers' questions were thought of positively. Robot's humorous comments added to its naturalness according to one person's observation. Another person mentioned that Nao's ability to give recommendations was an advantage.

**Questionnaire Results.** The demographic of the questionnaire responders was as follows: 50% female; 80.8% had Polish as their mother tongue; broad age range (14-58 yrs.; $M = 30.88$, $SD = 13.09$).

The means and standard deviations of the questionnaire results ($N = 26$) are shown in Table 2. All of them were on a scale 1 - 5, with 1 being "*strongly disagree*" and 5 being "*strongly agree*".

People were moderately interested in the interaction with robots with their reactions to the statement "*I think a robot could be a communication partner*" ($M = 3.46$, $SD = 1.24$) or "*I want to converse with a robot*" ($M = 3.12$, $SD = 1.42$). Surprisingly they expressed strong fear towards robots ($M = 4.52$, $SD = 1.04$) and some discontent with the idea of a robot's presence in their favorite cafe ($M = 3.00$, $SD = 1.60$). On the other hand, it seems as though they thought that generally robots could bean assistive ($M = 3.31$, $SD = 1.40$) entertaining ($M = 3.92$, $SD = 1.13$) in a cafe. Appearance-wise the participants thought that the robot's design should be pleasant ($M = 3.23$, $SD = 1.24$), yet it could be that they want it to have an original appearance, which would not be too human- ($M = 2.38$, $SD = 1.27$) or animal-like ($M = 1.88$, $SD = 1.10$). They were not familiar with similar devices ($M = 1.39$, $SD = 0.64$), and they thought other countries have already adapted robots compared to Poland ($M = 3.06$, $SD = 1.31$).

# 6 General discussion

## 6.1 Attitudes of participants Towards service robots in a cafe

Our study participants demonstrated a variety of behaviors related to the acceptance of and attitudes toward a social robot assistant in a cafe. We identified responses such as uncertainty, anxiety, mistrust, and fear towards the robot, which have been found in other studies [23, 15] as well. These may be related to the novelty factor and a lack of previous experience interacting with a robot [27]. Customers might not know the why the robot is in the cafe, whether it would understand them, or if it was trained to take their order. This may be related to the robot's appearance whose capabilities and body structure are not suitable for a cafe. Customers keeping physical distance from the robot may be interpreted as confirming a lack of trust toward the robot [25]. Some customers' negative attitudes toward the robot may be related to a generally bad attitude toward technology in public places among Poles [3]; it might be due to privacy concerns [28]. However, this attitude can be changed by increasing interaction facilities such as talking with the robot more naturally, thereby reducing unpleasant feelings such as fear, or stress in older customers or customers with little experience in using robots [29].

### 6.2    Factors in human-robot interaction affecting customer satisfaction

**First impressions:** Our study confirmed that the first impressions of the interaction play a crucial role in how it unfolds later [30]. We also assume that the appearance of the robot influences users' attitudes and it is crucial to meet customer expectations toward the robot design to provide positive feedback [31]. Our study confirmed a slight preference for human-like or animal-like appearance [18]. Moreover, the childlike look was perceived by some people as a positive feature that encouraged them to start conversation with the robot. Also, when the robot is pleasant, the customers perceive it positively and are more tolerant of its failures [32]. This can be related to the anthropomorphisation of the robot [33], which can positively influence human-robot interaction.  However, making the robot more human-like may trigger the "uncanny valley" effect [34], making the customer shy away from the conversation [35].

**Conversation flow:** Previous studies have shown that a proactive and responsive approach of the robot at the beginning is needed for the interaction to successfully proceed [36], and our results confirm this. We have observed people reluctant to start the conversation with questions, probably not sure whether the robot is going to answer; yet happily answering the ones asked by the robot. Furthermore, having a clearly assigned role of taking the customer orders, the robot convinced more people to talk to it, rather than when simply appealing as entertainment.

Several people were not willing to join the interaction. We presume, with some confirmation in our results, they did not know how to act, as it was a new situation for them. Such new situations require user-friendly guidance via the interaction [23, 15].

When users have too high expectations about the robot, they will be disappointed with its service quality and stop the interaction. However, if the robot's performance exceeds users' expectations, customers will be more satisfied with the provided service and will continue to use it [37].

Robot failures in understanding because of loud surroundings or slow responses affect the trust of people: how well a machine is regarded to carry out its task has a significant impact on how much trust people place in it [38, 15]. This suggests that machine mistakes are very likely to influence trust. On the other hand, it could suggest that customers may have had insecurity about their own ability to interact with technology, or that they were not used to interacting with robots.

**Adaptation to specific contexts:** The results show that recommendations were positively received by customers, as in the previous studies [17, 20, 39].

Our study revealed that people enjoyed the robot's clever and adequate answers. Its ability to conduct small talk was appreciated as well. The positive outcome corresponds with the fact that small talk, when desired, builds trust in interaction partners [40, 15] and with more trust comes better overall experience of the interaction [41].

People who ended up having a brief conversation with Nao were pleased by the interaction most of the time. Although, for some, the presence of the robot might have caused some confusion, we believe that the robot giving into the cafe scenario might have reduced part of that confusion by clearly indicating its function in the cafe sharing a task-oriented interaction with the human [42].

## 7    Conclusions and future research

Our goal was to study Polish consumers' attitudes towards assistive social robots in a public place, and to identify interaction features important for the customers, which could lead to a positive user experience. Despite the growth of robot assistants in Poland, they are still not widely available in public spaces like shops or cafes. Many people are familiar with assistive robots but service robots with human features are still alien to them. This might be a reason behind the customers' neutral attitude towards the Nao robot in a cafe. Even so, it is possible that these neutral emotions could be transformed into positive ones by familiarizing Polish society with such technologies. In future, we expect to observe customers' behavior and their adaptation to robots. After all, with digitalization in Poland came acceptance of various changes in everyday life and official realms. A similar change could happen with attitudes towards robot assistants in public places.

For a cafe scenario, the right design of a robot's interaction is important. We distinguished three essential aspects that affect customers' attitudes: *first impression*, *conversation flow*, and *adaptation* to a given context. Firstly, the appearance, voice and motor behavior of the robot should be pleasant for the users – from pleasant appearance and moderately human voice to appropriate gestures. It is effective if the robot initiates the interaction like a salesperson in a cafe would. It is also important that the robot shows its humanlike abilities including being as fluent as possible during the conversation. People react especially enthusiastically when the robot gives situation-specific comments and jokes. The robot should also accommodate peoples' willingness to have small talk while getting an order from the customer.

There were some limitations in our study. 1) The study lacked consistent counts of the participants, which can easily be avoided in the future. 2) The number of participants was not large enough. However, this is consistent with the contemporary approach to design which rejects universal solutions in favor of localized, culture-specific solutions [43]. Future studies on this topic should conduct more such studies with different user groups: for example, in rural areas and urban areas.3) As some of the robot's responses were provided on the spot and typed into a text-to-speech engine, they might have been too slow. Indeed, a few participants reported disengaging from the conversation due to delayed answers. We suppose that people's attitude could have been more positive towards the robot if the interaction was timely, thus the pre-designed conversation flow requires revision. 4) The robot's role might not have been clear to the customers, which is why they avoided any conversation in the first place. 5), Noise in the cafes created some problems. As our setup did not have any microphone to transfer the sound, it was difficult for the operator to hear what the customers were saying while sitting in another part of the cafe. In future studies, a microphone should be attached to the robot to make it easier to conduct the interaction.

Finally, we suggest that the future research on social robots in cafes should focus on designing the optimal conversation flow as it was revealed to be the most crucial aspect of customer satisfaction with such robots. Programming an autonomous robot with the right framework for conversation could result in automatizing the role of a cafe's salesperson with a satisfactory experience for the customers.